\documentclass{IEEEtran}
\usepackage{amsmath,amssymb,amsfonts}
\usepackage{algorithmic}
\usepackage{graphicx}
\usepackage{textcomp}
\usepackage{booktabs}
\usepackage{ragged2e}
\usepackage{array}
\usepackage{multirow}
\usepackage{makecell}
\usepackage{wrapfig}
\usepackage[numbers]{natbib}
\usepackage{longtable} 
\usepackage{tabularx,colortbl}
\usepackage{threeparttable}
\usepackage[breaklinks]{hyperref}
\usepackage[utf8]{inputenc}
\usepackage{adjustbox}
\def\BibTeX{{\rm B\kern-.05em{\sc i\kern-.025em b}\kern-.08em
    T\kern-.1667em\lower.7ex\hbox{E}\kern-.125emX}}
\begin{document}
\title{Cross-Task Generalization in Handwriting-Based Alzheimer’s Screening via Vision–Language Adaptation
}
\author{Changqing Gong, Huafeng Qin and Mounîm A. El-Yacoubi

\thanks{Changqing Gong and Mounîm A. El-Yacoubi are with Telecom SudParis, Institut Polytechnique de Paris, 91120 Palaiseau, France (e-mail: changqing.gong@telecom-sudparis.eu; mounim.el\_yacoubi@telecom-sudparis.eu).}

\thanks{Huafeng Qin is with the School of Computer Science and Information Engineering, Chongqing Technology and Business University, Chongqing 400067, China (e-mail: qinhuafengfeng@163.com).}

\thanks{Manuscript received xx; revised xx; accepted xx.
This work was supported in part by xxx, and in part by xxx.We also would like to thank Telecom SudParis and Institut Polytechnique de Paris for their partial support (Corresponding author: Huafeng Qin, Mounîm A. El-Yacoubi.)}
}

\maketitle
\begin{abstract}
Alzheimer's disease (AD) is a prevalent neurodegenerative disorder for which early detection is critical. Handwriting, which can be disrupted by subtle motor and cognitive decline, provides a non-invasive and cost-effective window for AD screening. Existing handwriting-based AD studies mostly rely on online trajectories and hand-crafted features, while the influence of handwriting task type on diagnostic performance and cross-task generalization remains underexplored. Meanwhile, large-scale vision--language models have demonstrated strong transfer and adaptation ability in natural-image anomaly detection and several medical modalities, such as chest X-ray and brain MRI. However, handwriting-based disease detection remains unexplored within this paradigm.
To address this gap, we introduce a lightweight Cross-Layer Fusion Adapter (CLFA) framework that repurposes Contrastive Language--Image Pre-training (CLIP) for handwriting-based AD screening. CLFA inserts multi-level adapters into a frozen visual encoder, combining cross-layer feature fusion with depthwise 2D convolution on patch grids to capture both local stroke irregularities and higher-level handwriting structure. This design progressively aligns pretrained vision--language representations with AD-related handwriting cues and supports transfer from supervised source tasks to task-disjoint unseen target tasks.
On the Darwin dataset, under the subject-disjoint cross-task protocol, averaged over all 600 task-disjoint source-target pairs, CLFA achieves 74.63\% AUC, 74.85\% accuracy, and 73.72\% F1 score, outperforming the best competing model by 2.15, 1.79, and 1.87 percentage points, respectively.

\end{abstract}


\section{Introduction}\label{Introduction}

Alzheimer's disease (AD), the most common cause of dementia, is a progressive neurodegenerative disorder characterized by cognitive decline affecting memory, reasoning, and daily functioning \cite{knopman2021alzheimer}. Early identification is increasingly important for timely clinical intervention and disease management \cite{long2019alzheimer(11),jack2024revised}. However, established biomarker modalities, such as amyloid positron emission tomography (PET) and cerebrospinal fluid (CSF) assays, remain constrained in large-scale screening by their cost, limited accessibility, and/or invasiveness \cite{hampel2018blood,karikari2022blood}. Neuropsychological assessments, including the Mini-Mental State Examination (MMSE) and Montreal Cognitive Assessment (MoCA), are more accessible but remain sensitive to demographic, educational, and administration-related factors \cite{tsoi2015cognitive,kourtis2019digital}. These limitations motivate the development of accessible, objective, and scalable digital approaches for AD screening.

Digital behavioral biomarkers have consequently received increasing attention, with studies exploring motor activity, speech, eye movements, gait, and digital cognitive behavior \cite{qi2025alzheimer}. Among these modalities, handwriting is particularly informative because it requires coordinated cognitive, visuospatial, linguistic, and fine motor processing \cite{fernandes2023handwriting}. Online handwriting further preserves the temporal evolution of the writing process, including trajectory, pressure, velocity, pauses, and in-air movements. Previous studies have reported alterations in handwriting timing, fluency, pressure, and movement dynamics in AD and mild cognitive impairment \cite{el2019aging(1),fernandes2023handwriting,cilia2024word}.

Computational approaches for handwriting-based AD detection have evolved from manually engineered kinematic and spatial descriptors \cite{cilia2022diagnosing} toward learned representations of handwriting dynamics. Sequential representation learning has been used to model temporal handwriting behavior \cite{el2019aging(1)}, while deep transfer learning and convolutional architectures have been employed to learn representations directly from handwriting signals or derived images \cite{cilia2021online,erdogmus2023promise}. More recently, hybrid Transformer architectures have integrated 1D dynamic signals with 2D handwriting images \cite{gong2025hybrid}, while stroke-level analysis has demonstrated the value of preserving fine-grained movement information beyond task-level statistical aggregation \cite{nardone2025handwriting}.

Despite these promising results, no works have examined how handwriting task affects both in-domain performance and cross-task generalization. Recently, large-scale  pre-trained visual-language
models (VLMs) have witnessed substantial advancements, applied to many different
scenarios \cite{ma2024deepcache,yu2023task,yang2023diffusion}, VLMs have delivered impressive zero/few-shot anomaly detection in natural images \cite{zhu2024toward,radford2021learning,goh2021multimodal,taori2020measuring}, and strong performance across several medical imaging modalities \cite{huang2024adapting} (e.g., chest X-ray, brain MRI, histopathology, retinal OCT). However, to our knowledge, handwriting-based disease detection remains unexplored within the VLMs paradigm. In particular, unseen-task cross-task generalization---training with labeled samples from source handwriting tasks and evaluating on task-disjoint unseen handwriting tasks---has never been addressed in handwriting-based AD screening, leaving open questions about task transferability and out-of-distribution robustness.
In this work, unseen-task cross-task generalization refers to training with labeled AD/healthy samples from source handwriting tasks and evaluating on task-disjoint target handwriting tasks that are not used during training. 
This setting should not be confused with unseen-disease-class zero-shot recognition, since the disease categories remain fixed as AD versus healthy control. As an example of cross-task generalization, we could consider a model trained on hand-drawn spirals and tested on handwritten words.

Actually, task transferability and out-of-distribution robustness of handwriting tasks, whether non-medical or medical, have never been  studied, leaving the potential of VLMs in this field unexplored.

To address this gap, we introduce Cross-Layer Fusion Adapters (CLFA), a lightweight multi-level adapter framework that repurposes the Contrastive Language-Image Pre-training (CLIP) model for handwriting-based AD detection.
At several depths of the ViT (Vision Transformer) image encoder, we implant residual fusion adapters that (i) project tokens into a compact bottleneck, (ii) reshape the patch (token) sequence into its 2D spatial grid and apply a depthwise 2D convolution to capture local neighborhood context, and (iii) fuse the current-layer descriptor with the already-fused descriptor propagated from the preceding adapter. The fused descriptor is then mapped back to the backbone width and injected as a small residual update, progressively retargeting the visual features from natural-image semantics to handwriting-specific medical cues while keeping the CLIP backbone intact. The fused mid-level descriptors at each tapped layer are treated as patch-level “detection tokens” and compared to normal/abnormal (AD-related) CLIP text prototypes, with multi-level aggregation yielding image-level anomaly scores for unseen-task screening. Our contributions are as follows:

\begin{itemize}
\item \textbf{}
We propose CLFA, a lightweight cross-layer adaptation framework for handwriting-based Alzheimer's disease screening.
CLFA progressively propagates and refines adapted descriptors across the frozen CLIP visual hierarchy.
At each selected layer, current-layer token representations are fused with previously adapted descriptors, followed by a lightweight residual update.
This design enables handwriting information to be progressively integrated across encoder depths while preserving the pretrained representations of the CLIP backbone.

\item \textbf{} 
To better capture fine-grained handwriting morphology, we introduce a spatial modeling operation within the adapter.
Patch tokens are restored to their spatial layout and processed using a depthwise 2D convolution, explicitly modeling local neighborhood interactions among adjacent handwriting regions.
This introduces a handwriting-relevant spatial inductive bias into the adapter, enabling CLFA to better represent local stroke structures and geometric variations.

\item \textbf{} 
We establish a cross-task evaluation protocol on the DARWIN benchmark. Experiments are conducted under both task-disjoint and subject-and-task-disjoint settings, providing a rigorous assessment of generalization to unseen handwriting tasks and unseen participants. The resulting cross-task evaluation offers a benchmark for studying task-robust and subject-independent Alzheimer's screening from handwriting.
\end{itemize}

Next, Section 2 reviews the state of the art. Section 3 outlines the DARWIN dataset tasks and data preprocessing. Section 4 details our proposed model and loss functions. Section 5 presents our experiments, comparing results with state-of-the-art models and analyzing the findings.

\section{Related Work}\label{sec:related}

Deterioration in writing ability is a recognized indicator of AD~\cite{garre2017kinematic(16)}, and kinematic handwriting analyses have revealed pathological patterns during the writing process~\cite{el2019aging(1)}.
Handwriting-based AD detection spans traditional machine learning (ML) and deep learning (DL).
Classical ML approaches extract temporal, kinematic, and spatial descriptors (e.g., speed, pressure, curvature) and train shallow classifiers such as logistic regression, SVMs, decision trees, or random forests~\cite{qi2023study,chai2023classification(34),meng2022image,cilia2022diagnosing}.
These methods often depend on extensive feature engineering and dataset-specific tuning.

DL has accelerated handwriting-based screening for neurodegenerative disorders, notably Parkinson’s Disease (PD) and, to a lesser extent, AD.
Several studies transform 1D pen signals into 2D representations and apply CNN-based models~\cite{pereira2018handwritten,taleb2023detection,diaz2021sequence, mwamsojo2022reservoir}.
For AD, \cite{cilia2022diagnosing} introduced the DARWIN dataset; subsequent works combine handcrafted or CNN-derived image features with ML heads~\cite{cilia2021online,cilia2022deep,erdogmus2023promise}. The work in ~\cite{dao2022detection} introduced 1D-CNNs with data augmentation for early-stage detection by modeling directly the online handwriting trajectory. In addition, 2D handwriting images were integrated with 1D feature signals \cite{gong2025hybrid}, enhancing the ability to capture the multimodal characteristics of handwriting.

Translating CLIP-based adaptation to medical anomaly detection is substantially more challenging due to large domain gaps across imaging modalities, cohorts, and acquisition protocols. Existing medical anomaly-detection methods often formulate the problem as one-class learning using normal samples only~\cite{bao2024bmad,jiang2023multi,zhang2020viral,zhou2020encoding,zhou2021proxy}. In contrast, CLIP learns aligned visual--textual representations through contrastive pre-training on large-scale image--caption pairs, allowing visual evidence to be compared with language descriptions in a shared embedding space. This property has motivated language-guided detection and segmentation methods, such as DenseCLIP and RegionCLIP~\cite{rao2022denseclip,zhong2022regionclip}, as well as multi-level visual adaptation for anomaly detection~\cite{huang2024adapting}.

More recent studies have further improved vision--language adaptation through biomedical prompt learning, knowledge-guided concept alignment, anomaly prompting, and adapter tuning. BiomedCoOp and vMFCoOp investigate prompt learning for biomedical vision--language models~\cite{koleilat2025biomedcoop,shao2026vmfcoop}, while Gao et al. align human knowledge with visual concepts to improve explainable medical image classification~\cite{gao2024aligning}. In the anomaly-detection setting, AnomalyCLIP learns object-agnostic normality and abnormality prompts~\cite{zhou2024anomalyclip}, and MMA introduces a multi-modal adapter for improving vision--language representation alignment~\cite{yang2024mma}. These works are closely related to our use of text-guided normal/abnormal prototypes and lightweight adaptation. However, they mainly address biomedical image classification, generic anomaly detection, or general VLMs adaptation. In contrast, our study focuses on handwriting-based Alzheimer’s disease screening under unseen-task cross-task generalization, where disease categories remain fixed as AD versus healthy control while handwriting task identities differ between training and testing.

Notably, no prior work has addressed cross-task generalization in handwriting-based AD screening, where a model is trained with labeled samples from source handwriting tasks and evaluated on unseen target tasks.

To address this gap, we repurpose a pre-trained vision–language model to build a novel handwriting AD anomaly detector and evaluate cross-task generalization, enabling an analysis of which handwriting factors most strongly relate to AD-linked anomalies. These gaps motivate our study: repurposing a vision--language pretrained model for handwriting-based AD, with an unseen-task cross-task generalization analysis. In Section IV, we will show how our proposed lightweight adapters can be integrated into CLIP to handle domain shift, enabling prompt-based Alzheimer’s screening from kinematic-derived 2D handwriting images.

\section{Material and method}\label{Material and method}
In this section, we introduce the dataset, describe our preprocessing of raw signal data, the extraction of 1D signal features and the reconstruction of handwriting images.
\subsection{Dataset}
We used the DARWIN-RAW dataset \cite{cilia2022diagnosing}, a widely used public benchmark for handwriting-based AD research, with data from 174 participants (89 AD patients and 85 healthy controls). This dataset includes 25 handwriting tasks designed for early AD detection \cite{cilia2018experimental(26)}, categorized into three types: memory and dictation (M), Copy (C), and Graphic (G).  This process was motivated by the effectiveness of kinematic features in detecting early AD. Table~\ref{tab:realname} lists the 25 handwriting tasks comprised in the Darwin dataset. 
\begin{figure*}[t]
\centerline{\includegraphics[scale=0.55]{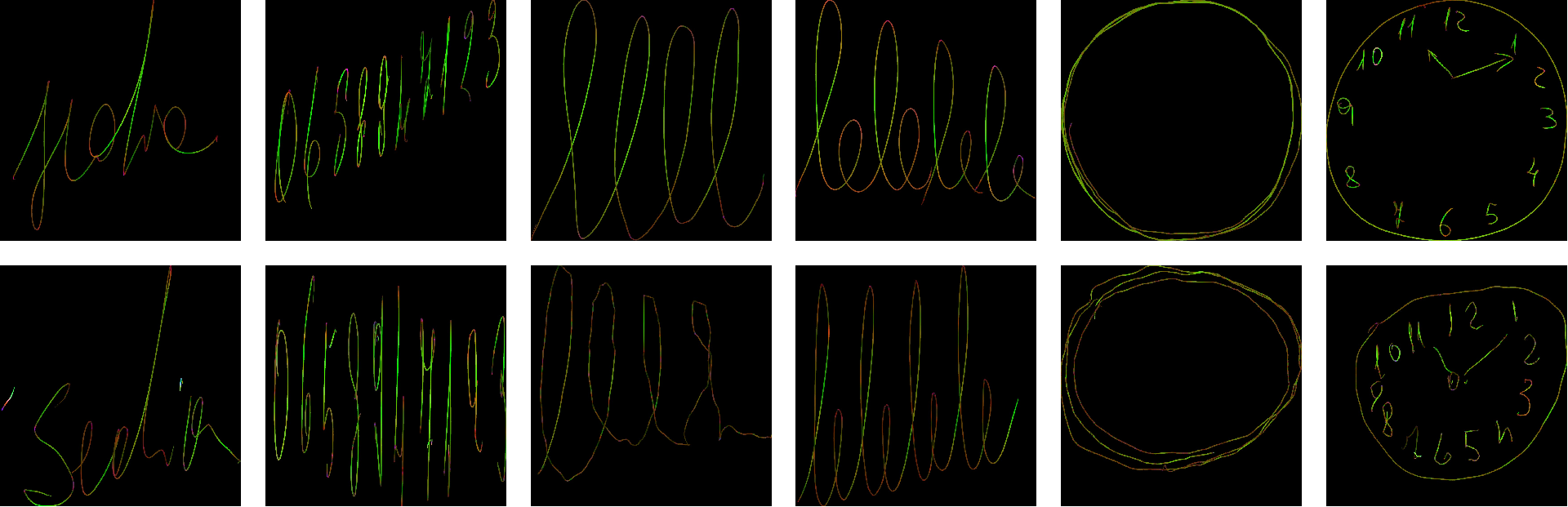}}
\caption{Example trajectory images generated from online handwriting. 
Top row: healthy controls; bottom row: AD patients. 
We show representative samples from the three major task categories: 
memory/dictation (M: Task~14, Task~22), 
copying (C: Task~8, Task~9), 
and graphic drawing (G: Task~4, Task~24). 
While some handwriting impairments in AD subjects are relatively easy to diagnose (e.g., distorted or trembled strokes), 
others remain subtle and difficult to distinguish from healthy controls, reflecting the variability of early-stage AD detection.}
\label{fig:p1}
\end{figure*}

\subsection{Generation of 2D Images from Online Handwriting}\label{sec:imggen}
The raw handwriting data \((x_i, y_i, p_i)\) were preprocessed to generate 2D images.
Our preprocessing and 1D feature extraction steps largely follow the protocol in our previous work ~\cite{gong2025hybrid}, 
including handling of missing values, interpolation, outlier removal, 
and the derivation of kinematic descriptors such as velocity, acceleration, jerk, curvature, and pressure. 
For details, we refer the reader to~\cite{gong2025hybrid}.

Building on these signals, we render 2D trajectory images by segmenting each task into 
\emph{paper} (pen-contact) streams. 
Trajectories are min--max normalized to a fixed canvas and rasterized with distance-adaptive interpolation. 
Our RGB channels encode distinct handwriting dynamics: curvature (R), velocity (G), and jerk (B), with stroke thickness modulated by pen pressure. In this paper, as a proof of concept, we use only on-paper handwriting as input, the inclusion of off-paper handwriting being straightforward. This process was motivated by the effectiveness of kinematic features in detecting early AD. Examples of generated images are shown in Figure~\ref{fig:p1}.

\subsection{Language Feature Formatting}
Motivated by prior work on prompt ensembling~\cite{huang2024adapting,zhou2024anomalyclip,gao2024aligning,jeong2023winclip}, we construct normal/abnormal (Alzheimer's Disease (AD)-related) text prototypes with a two-layer scheme to reduce prompt sensitivity. As shown in Table~\ref{tab:realname}, we provide a concise description of the 25 handwriting tasks by simplifying the task definitions in \cite{cilia2022diagnosing}. To obtain natural-language descriptions and task abbreviations, we further condense the wording; as summarized in Table~\ref{tab:state-level}, we employ a compact library of handwriting-quality descriptors for the two states (normal, abnormal), rather than a single generic sentence. Different from generic natural-image prompts, our templates are designed to describe handwriting trajectory images, such as rendered online handwriting samples and pen-trajectory patterns.

Each state-level pattern contains a slot that is instantiated with a task-specific noun phrase, yielding phrases such as `clear signature' . At the template layer, we retain the generic CLIP photo templates for prompt ensembling and further construct handwriting-specific templates tailored to trajectory-rendered and tablet-captured handwriting samples, as summarized in Table~\ref{tab:template-level}.
Finally, for task $t$, the normal/abnormal prototypes are collected as
\begin{equation}
\mathbf{E}_t=\big[\mathbf{e}^{\text{nor}}_t,\ \mathbf{e}^{\text{abn}}_t\big]\in\mathbb{R}^{D\times 2}.
\label{eq:text-prototypes}
\end{equation}

\begin{table}[t]
\centering
\caption{Text prompt task names}
\label{tab:realname}
\begin{tabularx}{\linewidth}{@{}cX cX@{}}
\toprule
\textbf{ID} & \textbf{Task name} & \textbf{ID} & \textbf{Task name} \\
\midrule
1 & signature & 2 & horizontal line drawing \\
3 & vertical line drawing & 4 & large circle drawing \\
5 & small circle drawing & 6 & copied letter `L' `M' `P' \\
7 & copied letters & 8 & cursive letter writing \\
9 & cursive bigram writing & 10 & copied word `foglio' \\
11 & copied word `foglio' on line & 12 & copied word `mamma' \\
13 & copied word `mamma' on line & 14 & written memory words \\
15 & reversed word `bottiglia' & 16 & reversed word `casa' \\
17 & copied multi-word phrases & 18 & written object name \\
19 & postal form copy & 20 & dictated sentence writing \\
21 & complex shape drawing & 22 & copied phone number \\
23 & dictated phone number writing & 24 & hand-drawn clock \\
25 & paragraph transcription &  &  \\
\bottomrule
\end{tabularx}
\end{table}

\begin{table}[t]
\centering
\caption{State-level prompts for normal and abnormal handwriting.}
\label{tab:state-level}
\begin{tabularx}{\linewidth}{X X}
\toprule
\textbf{Normal} & \textbf{Abnormal} \\
\hline
[o] & distorted [o] \\
clear [o] & unclear [o] \\
well-written [o] & tremulous [o] \\
legible [o] & impaired [o] \\
neatly drawn [o] & [o] with shaky lines \\

normal [o] & abnormal [o] \\
healthy [o] & poorly written [o] \\
smooth [o] & irregular [o] \\
regular [o] & fragmented [o] \\
well-aligned [o] & poorly aligned [o] \\
\bottomrule
\end{tabularx}

\vspace{2pt}
\footnotesize{\textit{Note:} [o] denotes the task name placeholder (see Table~\ref{tab:realname}).}
\end{table}

\begin{table}[t]
\centering
\caption{Template-level prompts for text feature ensembling.}
\label{tab:template-level}
\begin{tabularx}{0.9\linewidth}{X}
\toprule
\multicolumn{1}{c}{\textbf{Templates}}\\
\midrule
a handwriting sample of [c]. \\
a pen-trajectory image of [c]. \\
a handwriting trajectory showing [c]. \\
a kinematic handwriting pattern of [c]. \\
a tablet-captured handwriting image of [c]. \\
an online handwriting sample of [c]. \\
a digital handwriting sample of [c]. \\
a handwriting image depicting [c]. \\
a trajectory-based handwriting sample of [c]. \\

a handwriting pattern corresponding to [c]. \\
a handwriting trajectory consistent with [c]. \\
a tablet-based handwriting sample of [c]. \\
a pen-based handwriting image of [c]. \\
a rendered handwriting image of [c]. \\
\bottomrule
\end{tabularx}

\vspace{2pt}
\footnotesize{\textit{Note:} [c] denotes the handwriting state label(see Table~\ref{tab:state-level}).}
\end{table}

\section{Proposed Work}\label{sec:proposed}

Handwriting-based AD screening requires models that can generalize across heterogeneous writing tasks while remaining lightweight at inference. 
To this end, we repurpose CLIP through a Cross-Layer Fusion Adapter (CLFA) framework that progressively adapts the pretrained vision backbone to handwriting-specific AD medical cues. 
Our CLFA model implants multi-level adapters within the visual encoder.
Each adapter performs tokenwise depthwise convolution and cross-layer fusion, yielding locally contextualized and hierarchically aligned representations suitable for AD screening under cross-task generalization.

\subsection{Problem Setup and Notation}\label{sec:setup}

We aim to repurpose a vision--language model trained on natural images, denoted $M_{\text{nat}}$, for image-level anomaly detection in handwriting, yielding an adapted model $M_{\text{ad}}$. Adaptation uses a handwriting training set $\mathcal{D}_{\text{train}}$ drawn from source writing tasks, while evaluation targets unseen tasks under a cross-task generalization protocol.

Let $\mathcal{D}_{\text{train}}=\{(x_i, c_i, t_i)\}_{i=1}^{K}$, where $x_i\in\mathbb{R}^{h\times w\times 3}$ is a handwriting image ($h$ is the height of the image, $w$ is the width), $c_i\in\{0,1\}$ is the image-level anomaly label ($0$ normal, $1$ abnormal(AD)), and $t_i\in\mathcal{T}_{\text{train}}$ is the task identity, with $K$ the number of training samples. The test set $\mathcal{D}_{\text{test}}$ is built from tasks $\mathcal{T}_{\text{test}}$ that are disjoint from training tasks, i.e., $\mathcal{T}_{\text{train}}\cap\mathcal{T}_{\text{test}}=\varnothing$.

For an input $x$, the visual backbone in $M_{\text{ad}}$ produces token sequences at selected layers $\ell\in\mathcal{L}$:
$\mathbf{X}^{(\ell)}\in\mathbb{R}^{N\times C}$ ($N$=classification Token+$S$ patches).
At each selected layer, our dual-head adapter outputs (i) a text-aligned patch descriptor and (ii) a residual update that is injected back into the backbone.

For each training task $t\in\mathcal{T}_{\text{train}}$, we precompute CLIP text prototypes
$\mathbf{E}_t=[\mathbf{e}^{\text{nor}}_t,\mathbf{e}^{\text{abn}}_t]\in\mathbb{R}^{D\times 2}$ via prompt ensembles. In training, the cross-layer fusion adapters leverage text-prompt prototypes for patch-level alignment and detection guidance. The pretrained visual backbone is kept fixed, and only the lightweight adapters are optimized to adapt CLIP to handwriting data.

Given a test image $x_{\text{test}}$ from an unseen task $t_{\text{test}}\in\mathcal{T}_{\text{test}}$, the model outputs an image-level anomaly score $s(x_{\text{test}})\in[0,1]$. 
Performance is reported under unseen-task transfer from $\mathcal{T}_{\text{train}}$ to $\mathcal{T}_{\text{test}}$, assessing robustness to previously unseen handwriting tasks and distributions.

\subsection{Cross-Layer Fusion Adapters}\label{sec:adapter}
In handwriting-based AD screening, fine-grained local structure and subtle spatial variations are often more informative than high-level natural-image semantics.
To this end, we introduce a cross-layer fusion mechanism that propagates mid-level cues across transformer blocks.
Each adapter refines the current-layer patch tokens by integrating a fused descriptor propagated from the preceding layer, while a depthwise 2D convolution over the reshaped 
${h_p\times w_p}$ patch grid injects lightweight local spatial inductive bias, where $h_p$ and $w_p$ denote the number of ViT patches along height and width.
This design enables efficient and modular adaptation, and produces mid-level descriptors that can be aggregated for AD screening.

\begin{figure*}[t]
\centerline{\includegraphics[scale=0.6]{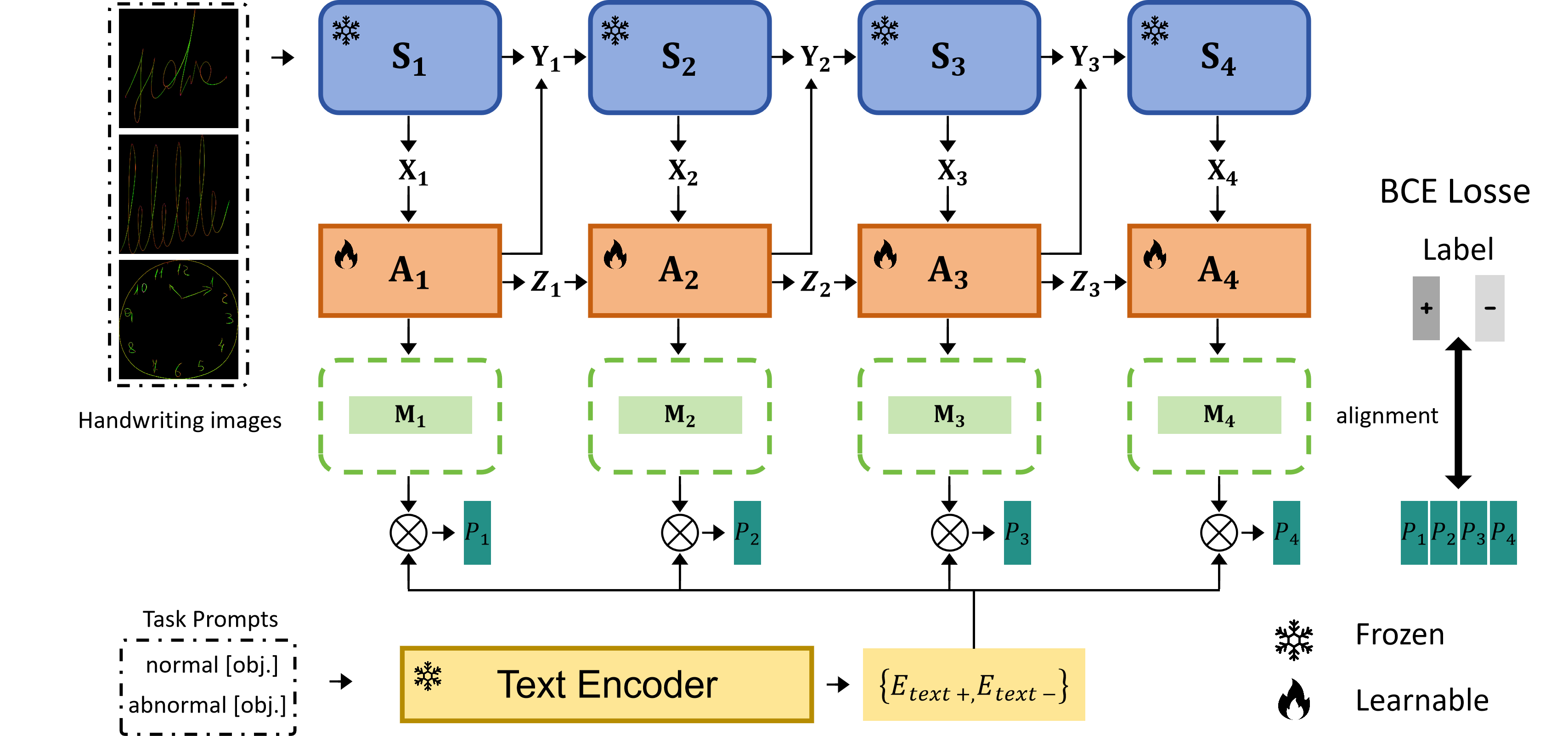}}
\caption{Overview of the Cross-Layer Fusion Adapter. 
Each selected ViT block hosts a lightweight adapter with depthwise 2D convolution and cross-layer fusion. 
Fused mid-level descriptors are compared to normal/abnormal CLIP text prototypes for AD screening under cross-task generalization.}
\label{fig:Frame}
\end{figure*}

Given an input image $x\!\in\!\mathbb{R}^{h\times w\times3}$, the CLIP visual encoder produces token features at selected layers $\ell\!\in\!\mathcal{L}$:
$\mathbf{X}^{(\ell)}\!\in\!\mathbb{R}^{N\times C}$, where $N$ denotes the number of tokens. 
At each selected layer, we implant a lightweight \emph{Fusion Adapter} that generates (i) a compact patch descriptor and (ii) a residual update softly injected back into the visual stream, as illustrated in Figure~\ref{fig:Frame}.

\paragraph{Adapter structure}
For layer $\ell$, the adapter first projects token features into a bottleneck space, as shown in Eq.\eqref{eq:proj}:
\begin{equation}
\mathbf{H}^{(\ell)} = \mathbf{X}^{(\ell)}\mathbf{W}_1,
\label{eq:proj}
\end{equation}
where $\mathbf{W}_1\!\in\!\mathbb{R}^{C\times T}$ and $\mathbf{H}^{(\ell)}\!\in\!\mathbb{R}^{N\times T}$. In our implementation, $T=256$, where $T$ denotes the adapter bottleneck dimension.
The projected patch tokens are reshaped into an $h_p{\times}w_p$ spatial grid and processed by a depthwise 2D convolution to capture local neighborhood context among nearby patches.
The output is then flattened back to a token sequence, as shown in Eq.~\eqref{eq:dwconv}:
\begin{equation}
\mathbf{M}'^{(\ell)} = \phi\!\big(\mathbf{H}^{(\ell)} + \mathrm{DWConv2D}(\mathbf{H}^{(\ell)})\big),
\label{eq:dwconv}
\end{equation}
where $\phi(\cdot)$ denotes the activation function. \(\mathrm{DWConv2D}(\cdot)\) operates only on the patch-token subset of \(\mathbf H^{(\ell)}\); the classification token bypasses the convolution and is concatenated back afterward.

\begin{figure}[ht]
\centerline{\includegraphics[width=0.65\columnwidth]{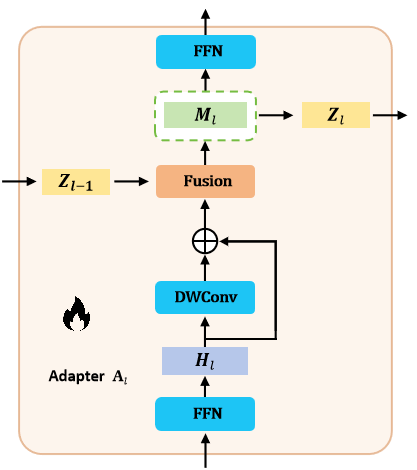}}
\caption{Cross-layer fusion Adapter. }
\label{fig:adapter}
\end{figure}

\paragraph{Cross-layer fusion.}
Let \(\mathcal{L}=\{\ell_1,\ell_2,\ldots,\ell_K\}\) denote the selected adapter layers, where \(K=4\) in our implementation. 
For the \(k\)-th selected layer \(\ell_k\), the depthwise-convolved descriptor is denoted as 
\(\mathbf{M}'^{(\ell_k)}\in\mathbb{R}^{N\times T}\). 
To propagate mid-level handwriting cues across different depths, we define a recurrent cross-layer fusion process:
\begin{equation}
\mathbf{Z}^{(\ell_k)} =
\begin{cases}
\mathbf{M}'^{(\ell_1)}, & k=1,\\[3pt]
\phi\!\left(
\mathbf{M}'^{(\ell_k)} + \mathbf{Z}^{(\ell_{k-1})}\mathbf{W}_{z}^{(k)}
\right), & k>1,
\end{cases}
\label{eq:fusion}
\end{equation}
where \(\mathbf{Z}^{(\ell_k)}\in\mathbb{R}^{N\times T}\) is the fused descriptor propagated from layer \(\ell_k\), 
\(\mathbf{W}_{z}^{(k)}\in\mathbb{R}^{T\times T}\) is a learnable projection that aligns the previous fused descriptor with the current adapter space, 
and \(\phi(\cdot)\) denotes the LeakyReLU activation. 
The fused descriptor \(\mathbf{Z}^{(\ell_k)}\) is then used as the cross-layer contextual input for the next selected adapter layer. 
This design allows local stroke-level evidence extracted at shallower layers to be progressively integrated with higher-level layout and shape representations.

\paragraph{Residual retargeting}
The fused descriptor is first mapped back to the original channel dimension through a lightweight projection, as shown in Eq.\eqref{eq:residual-proj}:
\begin{equation}
\mathbf{Y}^{(\ell)} = \phi\!\big(\mathbf{Z}^{(\ell)}\mathbf{W}_2\big),
\label{eq:residual-proj}
\end{equation}
where $\mathbf{W}_2\!\in\!\mathbb{R}^{T\times C}$ and $\phi(\cdot)$ denotes the activation function. 
The resulting output is then softly blended into the visual stream with a small residual coefficient:
\begin{equation}
\mathbf{X}^{(\ell)}_{\text{out}} = \,\mathbf{X}^{(\ell)} + \alpha\,\mathbf{Y}^{(\ell)},
\label{eq:residual-blend}
\end{equation}
where $\alpha{=}0.1$ in practice. 
As shown in Eq.~\eqref{eq:residual-blend}, this residual injection gently steers the pretrained CLIP backbone toward handwriting-specific evidence without disrupting global semantics.

\paragraph{Prototype matching and aggregation}
At each adapter depth, the fused  descriptors are first linearly projected to the CLIP text-embedding space with dimension $D$, then $\ell_2$-normalized and compared with the normal/abnormal CLIP text prototypes $\mathbf{E}_t\!\in\!\mathbb{R}^{D\times2}$, as shown in Eq. \eqref{eq:proto-match} :

\begin{equation}
\mathbf{A}^{(\ell)}=\mathrm{softmax}(\mathrm{norm}({\mathbf{Z}}^{(\ell)})\mathbf{E}_t)\in\mathbb{R}^{N\times2}.
\label{eq:proto-match}
\end{equation}
The abnormality probability for layer $\ell$ is obtained by mean pooling the abnormal channel, as shown in Eq. \eqref{eq:pool}:
\begin{equation}
p^{(\ell)}=\mathrm{Pool}\big(\mathbf{A}^{(\ell)}[:,1]\big),
\label{eq:pool}
\end{equation}

\paragraph{Detection objective}
The detection loss averages binary cross-entropy (BCE) over all selected layers, as shown in Eq. \eqref{eq:bce}:
\begin{equation}
\mathcal{L}_{\text{det}} = \frac{1}{|\mathcal{L}|} \sum_{\ell\in\mathcal{L}} 
\mathrm{BCE}\!\big(p^{(\ell)},\,y\big),
\label{eq:bce}
\end{equation}
where $y\!\in\!\{0,1\}$ is the ground-truth label (normal vs.~abnormal).

% \begin{figure*}[t]
%     \centering
%     \includegraphics[scale=0.6]{figs/auc_matrix_CLFA.png}
%     \caption{Cross-task AUC matrix for unseen-task AD screening using CLFA.
%     Rows denote training tasks; columns denote testing tasks. 
%     Higher off-diagonal values indicate stronger cross-task generalization.}
%     \label{fig:auc_matrix_CLFA}
% \end{figure*}

% \begin{figure*}[t]
%     \centering
%     \includegraphics[scale=0.6]{figs/auc_matrix_MVFA.png}
%     \caption{Cross-task AUC matrix for unseen-task cross-task AD screening using MVFA. 
%     Rows denote training tasks; columns denote testing tasks. 
%     Higher off-diagonal values indicate stronger cross-task generalization.}
%     \label{fig:auc_matrix_MVFA}
% \end{figure*}

\section{Experimental Results and Discussion}
In this section, we analyze the cross-task generalization behavior of handwriting-based Alzheimer’s screening using our CLFA framework. 
Rather than focusing solely on absolute detection performance, we aim to understand how handwriting task types influence unseen-task generalization and which task patterns best reveal Alzheimer-related handwriting impairments.

\subsection{Experimental Setup}
To assess our approach, we conducted extensive experiments on the DARWIN-RAW publicly available gold-standard dataset \cite{cilia2018experimental(26),cilia2022diagnosing,cilia2021online}, collected using Wacom's Bamboo tablet from 174 participants. The x-y coordinate sequences of pen-tip movements were recorded at a frequency of 200 Hz. The full dataset consists of x-y coordinates (174 subjects \( \times \) 25 tasks \( \times \) 1 x-y coordinate sequence, with some missing data). The x-y coordinates were then processed following the procedures described in Section 3. For a fair comparison, all CLIP-based methods are initialized from the same OpenAI-pretrained CLIP ViT-L/14 backbone and use the same input resolution of $240\times240$. The visual encoder contains 24 Transformer layers, which are divided into four stages of six layers each. During training, we use the AdamW optimizer with an initial learning rate of $1\times10^{-4}$, a weight decay of $3\times10^{-4}$, a batch size of 1, and a maximum of 50 epochs. All experiments are implemented in PyTorch and conducted on NVIDIA\texttrademark{} GPUs. We compare CLFA with several CLIP-based adaptation methods under consistent backbone and input settings, as well as representative handwriting-based classifiers initialized from pretrained models. Beyond conventional evaluation, we investigate cross-task generalization by training on one handwriting task and evaluating on unseen tasks.

\subsection{Adapter Architecture Configuration}
\label{sec:config}

The main architectural configurations of our CLFA model are summarized in Table~\ref{tab:model_config}.
For all CLIP-based experiments, we use a single backbone configuration to avoid architectural ambiguity: CLIP ViT-L/14 with input images resized to $(240\times240)$. The ViT-L/14 visual encoder contains 24 Transformer blocks. Following the multi-level design of MVFA for fair comparison, we divide the visual encoder into four stages and insert CLFA adapters after layers 6, 12, 18, and 24. The CLIP backbone is frozen and only the adapter parameters are optimized. All CLIP-based variants are evaluated under the same input resolution, data split, and task-transfer protocol.
Each adapter projects the intermediate token embeddings into a 256-dimensional adapter space, applies a tokenwise depthwise 2D convolution for local context modeling, fuses cross-layer descriptors, and re-injects the adapted features into the main stream via a small residual coefficient ($\alpha{=}0.1$).

\begin{table}[ht]
\centering
\caption{CLFA architectural settings.}
\label{tab:model_config}
\begin{tabularx}{0.9\linewidth}{>{\raggedright\arraybackslash}X >{\raggedright\arraybackslash}X}
\toprule
\textbf{Parameter} & \textbf{Value} \\
\midrule
Visual Backbone & CLIP ViT-L/14 \\
Input Resolution & $240\times240$ \\
Adapter Layers & [6, 12, 18, 24] \\
Adapter Dimension & 256 \\
Kernel Size & 3 \\
Residual Coefficient & 0.1 \\
\bottomrule
\end{tabularx}
\end{table}

\subsection{Cross-Task Generalization Analysis}

We first analyze the task-disjoint cross-task setting. In this experiment, the model is trained on one handwriting task and evaluated on different unseen handwriting tasks. This setting is used to characterize task-to-task transfer behavior and to examine how handwriting task type affects cross-task generalization. Unlike the subject-disjoint protocol introduced in the next subsection, this analysis focuses on task disjointness and does not explicitly enforce subject-level separation.

\begin{figure*}[ht]
    \centering
    \includegraphics[width=0.85\textwidth]{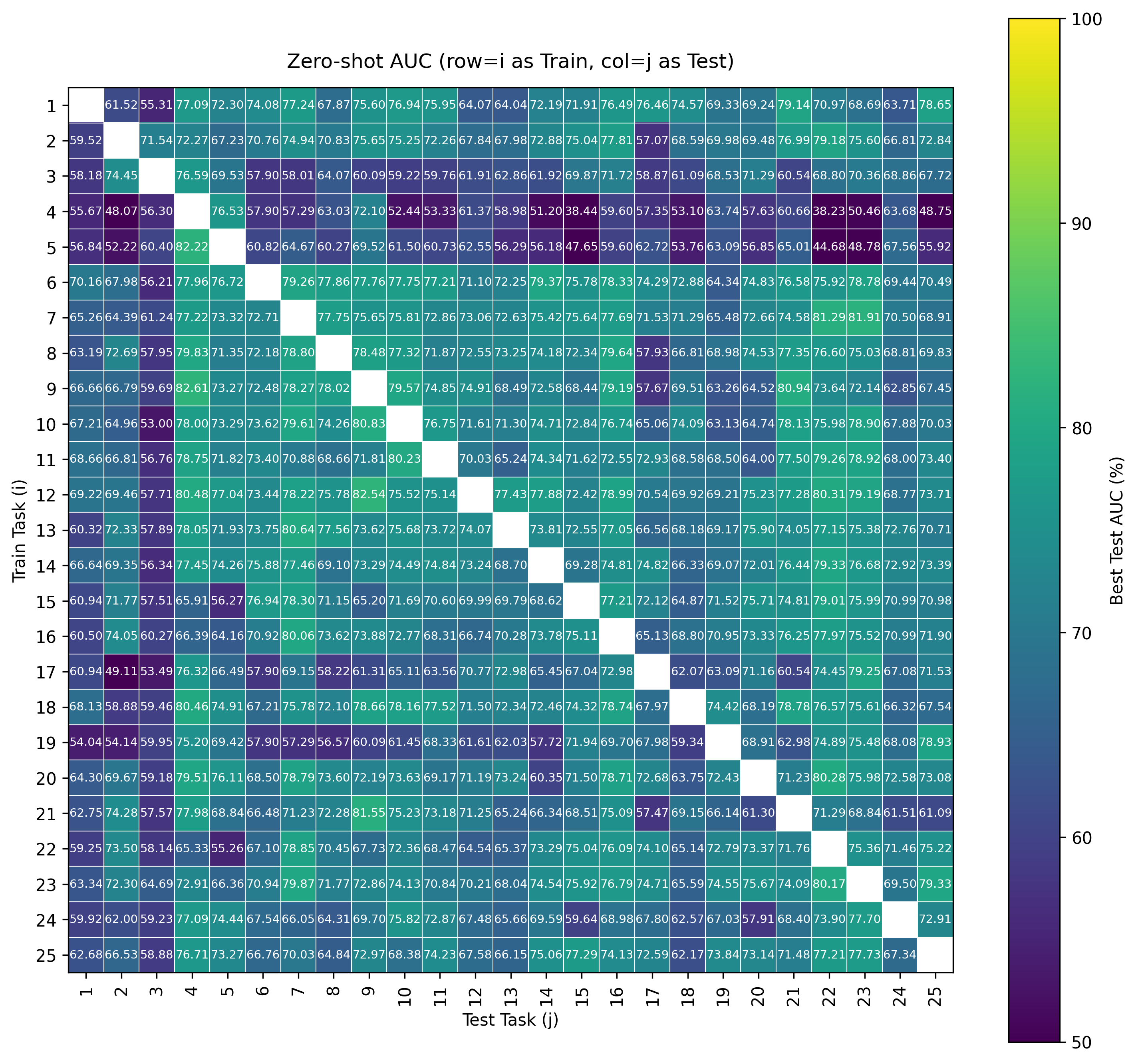}
    \caption{Cross-task AUC matrix of CLFA. Rows denote source tasks used for training, columns denote unseen target tasks used for testing, and each cell reports the AUC. }
    \label{fig:auc_matrix_CLFA}
\end{figure*}

Given a source handwriting task $s$, CLFA is trained using labeled samples from task $s$ and evaluated on each unseen target task $t$, where $t \neq s$. Each entry in the cross-task AUC matrix therefore represents one source-to-target transfer route. This matrix allows us to analyze two complementary aspects: source-task transferability and target-task detectability. Figure~\ref{fig:auc_matrix_CLFA} visualizes the full task-disjoint cross-task AUC matrix of CLFA. This analysis provides a task-level view of how handwriting representations learned from one task can transfer to other unseen tasks. To summarize the matrix, we compute the row mean and column mean. A higher row mean indicates that the source task provides more transferable representations, whereas a higher column mean indicates that the target task is more consistently detectable across different source tasks.

\begin{table}[ht]
\centering
\caption{Top-5 source and target tasks ranked by mean AUC in Cross-task generalization analysis.}

\label{tab:top_task_transfer_task_disjoint}
\begin{tabularx}{0.9\linewidth}{@{}>{\centering\arraybackslash}X >{\centering\arraybackslash}X >{\centering\arraybackslash}X@{}}
\toprule
Type & Task & Mean AUC (\%) \\
\midrule
\multirow{5}{*}{Source} 
& T12 & 74.39 \\
& T6  & 73.89 \\
& T7  & 72.87 \\
& T13 & 72.62 \\
& T23 & 72.46 \\
\midrule
\multirow{5}{*}{Target} 
& T4  & 76.35 \\
& T16 & 74.53 \\
& T23 & 73.68 \\
& T22 & 73.63 \\
& T7  & 73.36 \\
\bottomrule
\end{tabularx}
\end{table}

\begin{figure}[htbp]
    \centering
    \includegraphics[width=0.9\columnwidth]{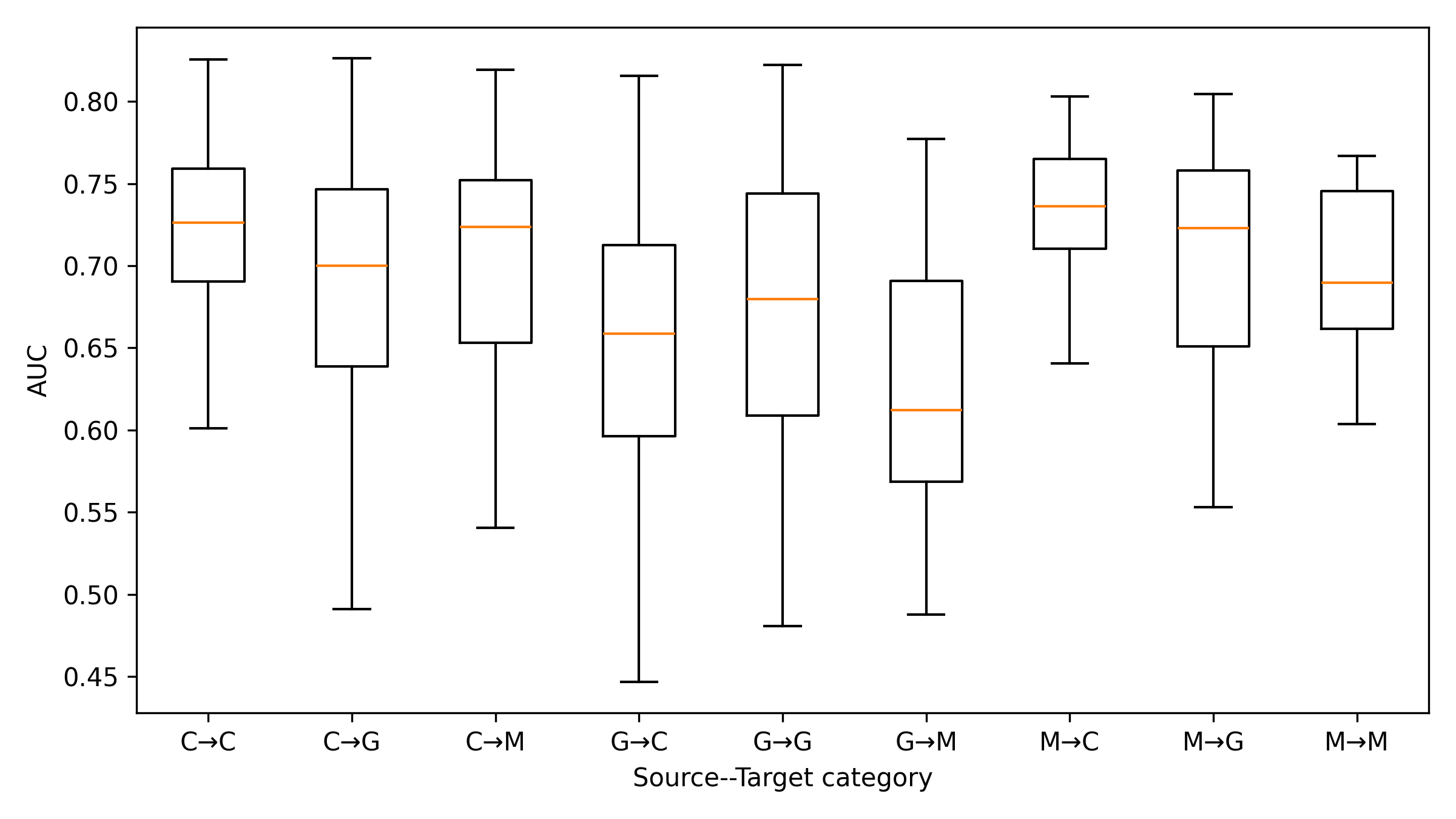}
    \caption{Cross-task AUC distributions across source–-target task categories.}
    \label{fig_box}
\end{figure}

\begin{table*}[ht]
\centering
\small
\setlength{\tabcolsep}{8pt} \renewcommand{\arraystretch}{1.15}
\caption{Subject-disjoint cross-task results over 600 source-target pairs, reported as mean $\pm$ standard deviation (\%).}
\label{tab:subject_disjoint_cross_task}
\begin{tabular}{lccccc}
\toprule
Method & AUC & Acc & F1 & Sens. & Spec. \\
\midrule
Vanilla CLIP & 51.35 $\pm$ 16.03 & 50.04 $\pm$ 6.50 & 37.67 $\pm$ 27.89 & 47.79 $\pm$ 41.44 & 52.30 $\pm$ 42.24 \\
WinCLIP & 57.38 $\pm$ 14.80 & 53.07 $\pm$ 7.20 & 50.99 $\pm$ 21.45 & 63.18 $\pm$ 36.42 & 42.97 $\pm$ 41.03 \\
Tip-Adapter & 59.91 $\pm$ 13.67 & 52.65 $\pm$ 7.01 & 54.01 $\pm$ 21.48 & 70.83 $\pm$ 36.02 & 34.48 $\pm$ 39.71 \\
CLIP-Adapter & 59.32 $\pm$ 13.98 & 52.46 $\pm$ 7.26 & 58.09 $\pm$ 16.50 & \textbf{76.45 $\pm$ 30.75} & 28.47 $\pm$ 35.13 \\
MVFA & 72.48 $\pm$ 9.27 & 73.06 $\pm$ 6.65 & 71.85 $\pm$ 9.27 & 72.02 $\pm$ 17.01 & 74.10 $\pm$ 16.99 \\

\midrule
\textbf{Ours} & \textbf{74.63 $\pm$ 7.51} & \textbf{74.85 $\pm$ 6.08} & \textbf{73.72 $\pm$ 8.25} & 73.51 $\pm$ 15.93 & \textbf{76.18 $\pm$ 15.41} \\
\bottomrule
\end{tabular}
\end{table*}

Table~\ref{tab:top_task_transfer_task_disjoint} summarizes the most transferable source tasks and the most detectable target tasks. Among source tasks, T12, T6, T7, T13, and T23 achieve the highest row-wise mean AUCs. These tasks mainly involve structured copying, word-level writing, letter production, or sequential number writing. Such tasks may provide relatively rich handwriting cues, including stroke continuity, inter-symbol spacing, baseline alignment, and sequential motor organization. These cues are not tied to a single target task and therefore appear more transferable across different handwriting activities. Among target tasks, T4, T16, T23, T22, and T7 show the highest column-wise mean AUCs. These tasks cover large-circle drawing, reversed-word writing, phone-number dictation, phone-number copying, and letter-copying tasks. Their high column-wise means suggest that AD-related handwriting patterns in these tasks can be detected from multiple source-task models. In particular, circle drawing may emphasize closure, curvature continuity, and spatial control; reversed-word writing and phone-number tasks may involve stronger sequencing and working-memory demands; and letter-copying tasks may reveal fine-grained stroke regularity and alignment. These task properties may make the corresponding target tasks more sensitive to handwriting abnormalities under task shifts. Figure~\ref{fig_box} summarizes cross-task performance across different task categories. The results show that relatively strong generalization occurs both within and across task categories, suggesting that CLFA captures AD-related handwriting information that remains useful under heterogeneous task conditions.

Overall, the task-disjoint analysis shows that cross-task generalization is not uniform across handwriting tasks. Some tasks are more effective as source tasks for learning transferable representations, whereas others are more sensitive as target tasks for detecting AD-related handwriting abnormalities. These results suggest that CLFA does not merely learn task-specific visual patterns, but captures handwriting cues that remain useful across different task types, including copying, graphic drawing, and memory/dictation tasks.

This task-level transfer behavior also motivates a more fine-grained research direction. While the present study evaluates AD-related handwriting cues at the image and task levels, the observed transfer patterns suggest that certain disease-relevant cues may be shared across heterogeneous handwriting activities. 

At the same time, this analysis should be interpreted cautiously. Although the observed transfer patterns are compatible with handwriting factors such as stroke continuity, spatial organization, symbol sequencing, and fine-motor stability, the dataset does not provide independent neuropsychological scores or clinician-annotated impairment mechanisms. Therefore, we interpret these findings as task-level transfer associations rather than direct evidence of specific underlying neurocognitive pathways.

\subsection{Subject-Disjoint Cross-Task Evaluation}

To reduce the risk of subject-specific handwriting leakage, we conduct a stricter subject-disjoint cross-task evaluation. In this protocol, all participants are first divided at the subject level into two non-overlapping cohorts, where 80\% of the subjects are used for training and validation, and the remaining 20\% are held out exclusively for testing. The split is performed before constructing any task-specific samples, ensuring that handwriting samples from the same participant never appear in both training and testing.

Given a source handwriting task $s$, the model is trained only on samples from task $s$ collected from the training subjects. It is then evaluated on each unseen target task $t$, where $t \neq s$, using only samples from the held-out testing subjects. Therefore, this protocol is both subject-disjoint and task-disjoint: no participant appears in both cohorts, and no target-task samples are used during model training.
For comparison, we include Vanilla CLIP~\cite{radford2021learning}, CLIP-Adapter~\cite{gao2024clipadapter}, Tip-Adapter~\cite{zhang2022tipadapter}, and WinCLIP~\cite{jeong2023winclip} as CLIP-based baselines. We also compare with MVFA~\cite{huang2024adapting}, a strong medical anomaly detection baseline. All methods are evaluated under the same subject-disjoint and task-disjoint protocol.

Table~\ref{tab:subject_disjoint_cross_task} summarizes the subject-disjoint cross-task performance over all source-target pairs, including 600 source-target pairs from 25 source tasks and 24 unseen target tasks per source task. The evaluation metrics include AUC, Accuracy, F1 score, Sensitivity, and Specificity. Direct CLIP-based baselines perform poorly on rendered handwriting trajectories. Vanilla CLIP obtains an average AUC of only 51.35\%, which is close to random performance. WinCLIP, Tip-Adapter, and CLIP-Adapter improve the AUC to 57.38\%, 59.91\%, and 59.32\%, respectively, but their accuracy remains close to 53\%. These results suggests that generic vision--language alignment alone is insufficient for robust handwriting-based Alzheimer's disease screening under task and subject shifts.

Compared with CLIP-based baselines, MVFA achieves substantially stronger generalization, with an average AUC of 72.48\% and accuracy of 73.06\%. Nevertheless, the proposed method further improves the average AUC to 74.63\% and the accuracy to 74.85\%. It also achieves the best F1 score and specificity among all methods. Compared with MVFA, our method improves AUC by 2.15 percentage points, accuracy by 1.79 percentage points, F1 score by 1.87 percentage points, and specificity by 2.08 percentage points.

The advantage of the proposed method can be attributed to its handwriting-oriented adaptation of CLIP. Instead of relying only on global visual--textual matching, CLFA progressively fuses multi-level visual representations and injects local spatial adaptation through depthwise 2D convolution on patch grids. This design helps preserve pretrained vision--language knowledge while enhancing sensitivity to fine-grained handwriting cues, such as stroke irregularity, spatial layout variation, and local trajectory structure. As a result, the model can better capture disease-related handwriting patterns that transfer across both unseen tasks and unseen subjects.

It is also worth noting that CLIP-Adapter achieves the highest sensitivity, but its specificity is much lower than that of our method. This indicates that CLIP-Adapter tends to over-identify AD samples at the cost of misclassifying healthy controls. In contrast, the proposed method provides a more balanced trade-off between sensitivity and specificity, which is important for clinically meaningful AD screening. Overall, the subject-disjoint results demonstrate that the proposed method does not rely on subject-specific handwriting memorization and generalizes more effectively to unseen participants and unseen handwriting tasks.

\begin{table}[ht]
\centering
\caption{
Comparison with existing handwriting-based classifiers. Results are reported as mean performance (\%).}
\label{tab:cross_task_generalization_auc_acc_f1}
\setlength{\tabcolsep}{2pt} \renewcommand{\arraystretch}{1.15}
\begin{tabularx}{0.95\linewidth}{@{}l >{\centering\arraybackslash}X >{\centering\arraybackslash}X >{\centering\arraybackslash}X >{\centering\arraybackslash}X >{\centering\arraybackslash}X >{\centering\arraybackslash}X >{\centering\arraybackslash}X >{\centering\arraybackslash}X >{\centering\arraybackslash}X@{}}
\toprule
\multirow{2}{*}{Model} 
& \multicolumn{3}{c}{T3} 
& \multicolumn{3}{c}{T11} 
& \multicolumn{3}{c}{T21} \\
\cmidrule(lr){2-4} \cmidrule(lr){5-7} \cmidrule(lr){8-10}
& AUC & Acc & F1 
& AUC & Acc & F1 
& AUC & Acc & F1 \\
\midrule
CNN-2D (AD)      
& 51.37 & 51.74 & 52.56 
& 51.41 & 50.00 & 58.95 
& 50.26 & 50.19 & 65.20 \\

MobileNetV2     
& 55.14 & 53.04 & 62.09 
& 55.90 & 53.59 & 64.01 
& 58.06 & 52.25 & 67.43 \\

SwinTransformer
& 63.58 & 54.17 & 63.93 
& 59.39 & 54.99 & 60.14 
& 57.57 & 53.59 & 61.47 \\

\midrule
\textbf{Ours}   
& \textbf{75.73} & \textbf{75.54} & \textbf{74.15} 
& \textbf{75.80} & \textbf{75.74} & \textbf{73.73} 
& \textbf{71.29} & \textbf{71.95} & \textbf{70.74} \\
\bottomrule
\end{tabularx}
\end{table}

\subsection{Comparison with existing handwriting-based classifiers}

To further evaluate whether conventional handwriting-based AD classifiers can generalize beyond the specific task used for training, we compare our method with representative image-based classifiers, including CNN-2D (AD)~\cite{erdogmus2023promise}, Swin Transformer~\cite{liu2021swin}, and MobileNetV2~\cite{sandler2018mobilenetv2}. Each model is trained on one representative source task, namely T3, T11, or T21, and evaluated on the remaining 24 unseen target tasks. The average AUC, Accuracy, and F1 score are reported in Table~\ref{tab:cross_task_generalization_auc_acc_f1}.

Overall, conventional handwriting classifiers show limited cross-task transferability. Although Swin-T achieves the best AUC among the competing baselines for T3 and T11, its performance remains substantially lower than that of our method. MobileNetV2 performs slightly better on T21 among the baselines, but still shows a clear gap compared with our approach. These results suggest that standard image classifiers tend to learn task-dependent visual patterns and struggle to capture disease-related handwriting cues that remain stable across different writing tasks.

In contrast, our method achieves the best performance under all three source-task settings. In terms of AUC, our method obtains 75.73\%, 75.80\%, and 71.29\% when trained on T3, T11, and T21, respectively. Compared with the strongest competing baseline for each source task, the corresponding AUC gains are +12.15, +16.41, and +13.23 percentage points. Similar improvements are also observed in accuracy and F1 score, indicating that the proposed CLFA learns more transferable handwriting representations rather than relying only on task-specific visual cues.

These results further support the importance of cross-task generalization in handwriting-based AD screening. They also show that simply applying conventional CNN or Transformer-based image classifiers is insufficient for robust unseen-task transfer, whereas the proposed cross-layer adaptation strategy is more effective for capturing task-invariant AD-related handwriting patterns.

\subsection{Ablation Study}

To verify the effectiveness of the two key components in CLFA, namely Cross-Layer Fusion (CLF) and depthwise 2D convolution (DW-2D), we conduct a controlled ablation study under the subject-disjoint cross-task setting. 
To reduce the influence of any single handwriting task and provide a more balanced evaluation, six representative source tasks are selected to cover different handwriting categories: T1, T3, T11, T15, T21, and T23. 
For each source task, the model is evaluated on the remaining 24 unseen target tasks, resulting in 144 task-disjoint source-target pairs for each variant.

We compare four variants: (i) w/o CLF \& DW-2D, where both proposed components are removed; (ii) w/o CLF, where only the cross-layer fusion module is removed; (iii) w/o DW-2D, where only the depthwise 2D patch-grid convolution is removed; and (iv) the full CLFA model. 
Table~\ref{tab:ablation_study} summarizes the average results over all 144 source-target pairs. 
Removing both CLF and DW-2D leads to the lowest performance, with an average AUC of 69.59\% and accuracy of 71.42\%. 
Introducing either component alone clearly improves the results. 
The w/o CLF variant achieves 72.25\% AUC and 73.34\% accuracy, while the w/o DW-2D variant achieves 72.36\% AUC and 73.19\% accuracy. 
These improvements show that both cross-layer fusion and local patch-grid adaptation contribute to cross-task transfer.

\begin{table}[ht]
\centering
\caption{Ablation study in the subject-disjoint cross-task setting. 
Results are averaged over six representative source tasks.}
\label{tab:ablation_study}
\setlength{\tabcolsep}{2pt} \renewcommand{\arraystretch}{1.15}
\begin{tabularx}{0.95\linewidth}{@{}l >{\centering\arraybackslash}X >{\centering\arraybackslash}X >{\centering\arraybackslash}X@{}}
\toprule
Variant & AUC & Acc & F1 \\
\midrule
w/o CLF \& DW-2D 
& 69.59 $\pm$ 9.16 
& 71.42 $\pm$ 6.07 
& 69.66 $\pm$ 9.65 \\

w/o CLF 
& 72.25 $\pm$ 8.34 
& 73.34 $\pm$ 5.99 
& 72.72 $\pm$ 7.89 \\

w/o DW-2D 
& 72.36 $\pm$ 9.20 
& 73.19 $\pm$ 6.67 
& 72.76 $\pm$ 8.52 \\

\textbf{Ours} 
& \textbf{74.61 $\pm$ 7.37} 
& \textbf{74.85 $\pm$ 6.08} 
& \textbf{73.50 $\pm$ 8.48} \\
\bottomrule
\end{tabularx}
\end{table}

The full CLFA achieves the best overall performance, reaching 74.61\% AUC, 74.85\% accuracy, and 73.50\% F1 score. 
Compared with the variant without both components, CLFA improves AUC by 5.02 percentage points and accuracy by 3.43 percentage points. 
Compared with the single-component variants, CLFA also provides consistent gains, indicating that CLF and DW-2D are complementary rather than redundant. 
Specifically, CLF helps integrate handwriting cues across different representational depths, while DW-2D strengthens local spatial modeling on the patch grid. 
Their combination enables the model to better capture both local stroke-level irregularities and higher-level handwriting structures.

\begin{table*}[ht]
\centering
\scriptsize
\caption{Per-source ablation results in the subject-disjoint cross-task setting. 
Each source task is evaluated on 24 unseen target tasks. 
Results are reported as mean performance (\%).}
\label{tab:ablation_per_source_wide}
\setlength{\tabcolsep}{3pt}
\resizebox{\textwidth}{!}{
\begin{tabular}{lccc ccc ccc ccc ccc ccc}
\toprule
\multirow{2}{*}{Model Variant}
& \multicolumn{3}{c}{T1}
& \multicolumn{3}{c}{T3}
& \multicolumn{3}{c}{T11}
& \multicolumn{3}{c}{T15}
& \multicolumn{3}{c}{T21}
& \multicolumn{3}{c}{T23} \\
\cmidrule(lr){2-4}
\cmidrule(lr){5-7}
\cmidrule(lr){8-10}
\cmidrule(lr){11-13}
\cmidrule(lr){14-16}
\cmidrule(lr){17-19}
& AUC & Acc & F1
& AUC & Acc & F1
& AUC & Acc & F1
& AUC & Acc & F1
& AUC & Acc & F1
& AUC & Acc & F1 \\
\midrule
w/o CLF \& DW-2D
& 70.35 & 71.24 & 70.57
& 69.55 & 71.65 & 71.47
& 69.45 & 71.00 & 66.22
& 71.64 & 73.49 & 72.76
& 61.24 & 66.32 & 61.84
& 75.34 & 75.13 & 75.12 \\

w/o CLF
& 73.89 & 74.02 & 73.07
& 69.38 & 70.83 & 70.64
& 73.97 & 74.12 & 73.23
& 71.86 & 73.85 & 73.86
& 68.22 & 71.15 & 70.06
& 76.19 & 75.52 & 75.49 \\

w/o DW-2D
& 70.96 & 71.38 & 71.85
& 72.14 & 73.06 & 71.89
& 72.86 & 73.50 & 74.04
& 72.20 & 73.57 & 72.06
& 69.72 & 71.76 & 71.62
& 76.30 & 75.94 & 75.09 \\

Ours
& 74.88 & 74.76 & 73.91
& 75.73 & 75.54 & 74.15
& 75.80 & 75.74 & 73.73
& 72.78 & 74.44 & 73.11
& 71.29 & 71.95 & 70.74
& 77.17 & 76.47 & 75.37 \\
\bottomrule
\end{tabular}
}
\end{table*}

Table~\ref{tab:ablation_per_source_wide} further reports the per-source results. 
Across all six representative source tasks, Ours achieves the highest AUC and accuracy. 
For motor-dominant tasks such as T1 and T23, the full model obtains strong AUCs of 74.88\% and 77.17\%, suggesting that the proposed adaptation is effective for handwriting patterns involving signature production and dictated number writing. 
For graphic tasks such as T3 and T21, Ours also achieves the best AUC and accuracy, showing its ability to transfer from structured line drawing and complex-form retracing to other unseen tasks. 
For copying-related cognitive tasks such as T11 and T15, the full model again achieves the highest AUC and accuracy, indicating that CLFA can capture transferable cues from word copying and reversed writing.

Overall, the ablation results demonstrate that both CLF and DW-2D are important for robust cross-task generalization. 
CLF provides hierarchical feature integration across layers, while DW-2D enhances local patch-grid adaptation. 
The full CLFA combines these two mechanisms and consistently improves AUC and accuracy across different source-task types, supporting the effectiveness of the proposed cross-layer and local adaptation design for handwriting-based AD screening.

\section{Conclusion}
In this study, we investigated handwriting-based Alzheimer's disease (AD) screening from a cross-task generalization perspective. Unlike previous studies that mainly focus on task-specific classification, we examined whether disease-related handwriting cues learned from one source task can transfer to unseen target tasks. To this end, we introduced Cross-Layer Fusion Adapter (CLFA), a lightweight vision--language adaptation framework that repurposes CLIP for handwriting-based AD screening. By inserting multi-level adapters into the frozen visual encoder, CLFA combines cross-layer feature fusion with depthwise 2D convolution on patch grids, allowing the model to capture both local stroke irregularities and higher-level handwriting structure.

Extensive experiments on the DARWIN dataset demonstrate the effectiveness of the proposed framework. In the task-disjoint setting, the full $25\times25$ cross-task analysis suggest that different handwriting tasks share transferable cues, enabling models trained on one task to provide useful evidence on other unseen tasks rather than being restricted to the same task.
We further conducted a stricter subject-disjoint cross-task evaluation, where training and testing subjects are separated and target tasks remain unseen during training. CLFA achieves the best overall AUC, accuracy, F1 score, and specificity compared with CLIP-based baselines and existing handwriting-based classifiers. Ablation studies further confirm the importance of cross-layer fusion and local patch-grid adaptation for robust cross-task transfer.

Despite its promising generalization ability, this study has several limitations.
First, although the dataset covers diverse task types, it remains limited in demographic representation and writing conditions, which may constrain population-level generalizability.
Second, CLFA currently models temporal dynamics implicitly on the rendered image representations through handcrafted encoding of velocity, jerk, and pressure. We could instead model explicitly the temporal dynamics of pen trajectories to let the model to automatically learn feature dynamics representations that are sensitive to micro-motor fluctuations and AD handwriting impairments.
Third, the current task-level analysis reveals transferable handwriting patterns but does not localize which specific trajectory fragments contribute most to AD discrimination.
Future work will extend toward few-shot personalization, multi-modal fusion (e.g., integrating handwriting with speech or eye-tracking), and more efficient prompt-context learning to enable more comprehensive cognitive assessment in real-world settings.

\bibliographystyle{IEEEtran}
\bibliography{cas-refs}

\end{document}